\RequirePackage{fix-cm}
\documentclass[smallcondensed]{svjour3}     
\smartqed  
\usepackage{graphicx}
\usepackage{makecell}
\usepackage{algorithm}
\usepackage{algorithmic}
\usepackage{amsmath}

\begin{document}

\title{Task planning and explanation with virtual actions
}


\author{Guowei Cui         \and
        Xiaoping Chen 
}


\institute{School of Computer Science and Technology, University of Science and Technology of China, Hefei 230026, China \\
              \email{cuigw@mail.ustc.edu.cn; xpchen@ustc.edu.cn}           
}

\date{Received: date / Accepted: date}

\maketitle

\begin{abstract}
One of the challenges of task planning is to find out what causes the planning failure and how to handle the failure intelligently. This paper shows how to achieve this. The idea is inspired by the connected graph: each verticle represents a set of compatible \textit{states}, and each edge represents an \textit{action}. For any given the initial states and goals, we construct virtual actions to ensure that we always get a plan via task planning. This paper shows how to introduce virtual action to extend action models to make the graph to be connected: i) explicitly defines static predicate (type, permanent properties, etc) or dynamic predicate (state); ii) constructs a full virtual action or a semi-virtual action for each state; iii) finds the cause of the planning failure through a progressive planning approach. The implementation was evaluated in three typical scenarios.

\end{abstract}

\section{Introduction}
\label{intro}

Action reasoning, which makes decisions about how to act in the world, requires symbolic representations of the robot’s environment and the actions the robot can perform, has been widely used for task planning and control in many service robot applications~\cite{chenDevelopingHighlevelCognitive2010,chenPlanningTaskorientedKnowledge2016,jiangOpenWorldReasoningService2019,savageSemanticReasoningService2019,hanheideRobotTaskPlanning2017,stenmarkConnectingNaturalLanguage2015,yangPEORLIntegratingSymbolic2018}. Some works have focused on dealing with uncertainty and insufficient information in the environment to get a better planning result. Puigbo et al.~\cite{puigboUsingCognitiveArchitecture2015} adopt Soar cognitive architecture~\cite{lairdSoarCognitiveArchitecture2012} to support understanding users' commands. Chen et al. \cite{chenPlanningTaskorientedKnowledge2016} acquire task-oriented knowledge by HRI and continuous sensing. Hanheide et al.~\cite{hanheideRobotTaskPlanning2017} extends an active visual object search method~\cite{aydemirActiveVisualObject2013} to explain failures by planning over explicitly modeled additional action effects and assumptive actions. Some work have focused on dealing with unseen objects~\cite{jiangOpenWorldReasoningService2019,miningerInteractivelyLearningStrategies2016}. Some focus on using open source knowledge to handle incomplete information~\cite{chenOpenKnowledgeEnabling2013,luUnderstandingUserInstructions2016}. Lim et al.~\cite{limOntologyBasedUnifiedRobot2011} present an ontology-based unified robot knowledge framework that integrates low-level data with high-level knowledge to enable reasoning to be performed when partial information is available. Using theory of intensions~\cite{blountTheoryIntentionsIntelligent2015}, which is based on scenario division, \cite{srivastavaTractabilityPlanningLoops2015} generates plan with loops to accommodate unknown information at planning time. In \cite{al-moadhenSemanticBasedTask2015}, a semantic knowledge base is proposed to derive and check implicit information during plan generation. Most of the work improves the efficiency of planning and probability of success, but rarely involves the handling of planning failures, such as explaining the reasons for planning failures, and whether the cause of failure can be determined is useful for further processing. Hanheide et al.~\cite{hanheideRobotTaskPlanning2017} explain failures by planning over explicitly modeled additional action effects and assumptive actions, the execution is based on predefined models and is primarily for execution failures.

Some work~\cite{oetschStepwiseDebuggingAnswerset2018,nardiIntegratedEnvironmentReasoning2013,dodaroDebuggingNongroundASP2019,cuteriDebuggingAnswerSet2019} about debugging logic programs can be used to address the reasons for planning failures, but it is also difficult to automatically identify which parts of it are of interest to the programmer. To get this information, we construct virtual actions for dynamic predicates in the action domain. Virtual actions have as few preconditions as possible and far greater costs than normal actions. Because virtual actions have fewer restrictions, so they can be called when there is no solution with normal actions. The virtual actions can be used to address what state transitions cannot be completed via normal actions.

Section~\ref{sec:motivation} explains our motivations for this work in detail. Section~\ref{sec:domain} presents how to design the action domain. Virtual actions are described in Section~\ref{sec:virtual_action} and the planning algorithm is described in Section~\ref{sec:framework}. We report on our case studies in Section~\ref{sec:cases} and give some conclusions in Section~\ref{sec:conclusion}.

\section{Motivations}\label{sec:motivation}

Even when all action models are constructed correctly, in some cases, the planner can't find any plan for some goals. The reasons for these failures may be varied, but classified, we classify them into four categories:
\begin{itemize}
  \item P1 (lack of action): A state can't be achieved by any action, aka, it isn't any effect of any action but a precondition of action or the goal.
  \item P2 (layout problem): Some preconditions (dynamic predicates, states) of actions and environment prevent actions from working. It may be the distribution of some instance's position or another state that makes the path impossible (for example, The key needed to open the room is locked in the room). 
  \item P3 (static predicate problem): Planner can't find path due to some static predicate of an action needed is not met.
\end{itemize}

These three categories reflect different scenarios, and we will describe how to find the cause of the failure in the following. We are concerned with application scenarios that conform to the following two assumptions:
\begin{itemize}
  \item Sufficient computing power: Our method is used to find failure reason after planning failure, and sufficient computing power can ensure that the problem is not finding a solution, not too late to find a solution.
  \item Limited number of steps: If there is a plan to achieve the goal, the number of steps of this plan is no greater than the predefined value.
\end{itemize}

For P1 and P2, we construct full virtual actions and semi-virtual actions, which have the minimum preconditions to complete state transitions, to try our best to get a plan, even so, it may be possible to P3, in which even if the virtual actions are included, the planner can not get any plan. For P3, we define the action related to the gaol to find out which conditions are not being met. The main ideas are described below.

\begin{enumerate}
  \item Dynamic and static predicates are defined explicitly, and any parameter of each dynamic predicate needs to be restricted by some static predicates. This design is to generate the corresponding virtual action for each dynamic predicate.
  \item Planning with full virtual actions, once get a plan, we know that the state transitions required by the plan cannot be completed because some actions are not defined in the action domain.
  \item Planning with semi-virtual actions, once get a plan, we know that the state transitions required by the plan cannot be completed because there are some conflicts between action models and the world.
  \item Planning with full virtual and semi-virtual actions, once get a plan, we know that the state transitions required by the plan cannot be completed because there are some conflicts between action models and the world, and some actions are not defined in the action domain.
  \item Planning with full virtual and semi-virtual actions, if we can't get any plan, we know that the action, which contains the goal as its effect, its preconditions contains some static predicates which can not be met.
\end{enumerate}

\section{Domain formulation}
\label{sec:domain}

The action domain description consists of (i) \textit{static predicates}, (ii) \textit{dynamic predicates} (states), and (iii) \textit{actions}. There is a key point to focus on: each state is limited by some static predicates, which are mainly to generate a full virtual action or semi-virtual action for achiving any \textit{state} transition.

\subsection{Static predicate}
Static predicates are templates for logical facts that can be classes (e.g. room, apple, water cup), features (e.g. red, with lid). All static predicates are designed to be time-independent that means each of them will stay the same during planning. Static predicates are defined as the following.

\texttt{\\
static \{ \\
\indent room(R).\\
\indent object(O).\\
\indent apple(A).\\
\}}

The predicate \textit{apple(A)} defines a template means \textit{A}, which will be a real instance when this template is used, is an apple.

\subsection{\textit{Dynamic predicate}}\label{sec:state}
Each dynamic predicate, aka state, is a template that defines the state that an instance is in. Each state is thought to be able to be changed by the robot's actions or other events.

\texttt{\\
dynamic \{ \\
\indent opened(A) : door(A).\\
\indent isNear(B) : human(B) | location(B).\\
\indent isPlaced(A,B) : object(A) \& location(B).\\
\}
}

As shown above, each \textit{state} consists of two parts: \textit{head} (left part), which defines the name and parameters, and \textit{requirements} (right part), which are \textit{static predicates} limit all parameters. The \textit{requirements} can be in many cases, which are connected by symbol \textbf{$($}, \textbf{$)$}, \textbf{$|$} and \textbf{$\&$}. In each case, the parameters must be limited by \textit{static predicates}.

\subsection{Action}
Actions are operator-schemas, which should be grounded/instantiated during execution. Each action has a \textit{name} (including variables that should be instantiated with object), \textit{preconditions} and \textit{effects}. The effects of actions could be also \textit{conditional}.

\texttt{\\
action \{ \\
\indent name: moveTo(L,R).\\
\indent precondition:\\
\indent \indent inRoom(R) \& (isLocated(L,R) | locInRoom(L,R)) \& not {isNear(L)}.\\
\indent effect:\\
\indent \indent isNear(L) \& ($\neg${isNear(L1)} $\leftarrow$ isNear(L1)).\\
\}
}

In the action above, symbol $\neg$ means \textit{classical negation}, $not$ means \textit{non-classical negation} and $\leftarrow$ means conditional (left part is effect and right part is conditions). Each parameter must be limited by \textit{states} or types. For a more detailed introduction of action, you can refer to our previous work \cite{cuiSemanticTaskPlanning2021}.

\section{Virtual action}\label{sec:virtual_action}

Even when all action models are constructed correctly, in some cases, the planner can't find any plan for some goals. As Sec.~\ref{sec:motivation} described, we group the reasons for failure into three categories: 1) lack of action; 2)layout problem; 3) static predicate problem. To solve these problems, we introduced virtual actions. Virtual actions are generated from predefined dynamic predicates that contain only the basic preconditions, and due to the relaxation of preconditions, the planner has a greater probability of finding a plan. When the plan contains a virtual action, we can know which specific conditions cannot be met based on the virtual action. A virtual action contains only one effect, which is its corresponding dynamic predicate. And it contains two preconditions: the requirements of its corresponding dynamic predicate, and the non-classical negation of its effect. If a dynamic predicate is the effect of some action, the virtual action generated from the dynamic predicate is called \textit{semi-virtual action}, or the virtual action is \textit{full virtual action}. Alg. \ref{alg:gen_virtual_action} shows how to generate the virtual actions. 

\begin{algorithm}
  \caption{Generate virtual actions}
  \label{alg:gen_virtual_action}
  \begin{algorithmic}[1]
      \REQUIRE Actions $ A $, dynamic predicates and their classical negation $P_d$
      \ENSURE full virtual actions $A_f$, semi-virtual actions $A_s$
      \STATE{$A_f \leftarrow \emptyset$, $A_s \leftarrow \emptyset$}
      \STATE{add $not \: s$ to $R$}
      \FOR{\textbf{each} $ p \in P_d $ }
        \STATE{$isSemiVirtual \leftarrow False$}
        \STATE{$s,R,flag_{neg} \leftarrow p$}
        \FOR{\textbf{each} $a \in A$}
          \IF{$p$ is an effect of $a$}
            \STATE{$isSemiVirtual \leftarrow True$}
            \STATE{\textbf{break}}
          \ENDIF
        \ENDFOR
        \IF{$isSemiVirtual$ is $True$}
          \IF {$flag_{neg}$ is $True$}
            \STATE{add $('semi\_d\_'s, R\cup\{s\}, \neg{s})$ to $A_s$}
          \ELSE
            \STATE{add $('semi\_e\_'s, R\cup\{\textbf{not} \: s\}, {s})$ to $A_s$}
          \ENDIF
        \ELSE
        \IF {$flag_{neg}$ is $True$}
          \STATE{add $('full\_d\_'s, R\cup\{s\}, \neg{s})$ to $A_f$}
          \ELSE
            \STATE{add $('full\_e\_'s, R\cup\{\textbf{not} \: s\}, {s})$ to $A_f$}
          \ENDIF
        \ENDIF
      \ENDFOR
  \end{algorithmic}
\end{algorithm}

Alg. \ref{alg:gen_virtual_action} needs the action models and the dynamic predicates and ultimately returns both full and semi-virtual actions. In the beginning, both full and half-imaginary actions are empty sets, then for any dynamic predicate, do the following:
\begin{enumerate}
  \item Divide the dynamic predicate into two parts: head($s$) and requirements($R$).
  \item Add the non-classical negation of $s$ to $R$. 
  \item Determine if the dynamic predicate is the effect of some action, and if so, perform step 4, otherwise perform step 5.
  \item Construct a semi-virtual action and add it into set $A_s$.
  \item Construct a full virtual action and add it into set $A_f$.
\end{enumerate}

\section{Progressive reasoning framework}\label{sec:framework}

\subsection{Planning}
Two points have to be considered in the new planning method. First, it should have no or small impact on the successful cases. To reduce the impact on the successful cases, the new method does no additional operations before the planning fails. If the planner successfully gets a plan for the goals based on current states, it will dispatch the plan and wait for the execution results. The planner decides whether to re-plan based on execution feedback and environmental changes. For details on handling these normal situations, please refer to our previous work \cite{cuiSemanticTaskPlanning2021}. Second, virtual actions cannot drown out normal actions. The virtual actions should be called as a last resort. We achieve this goal by setting different costs for virtual and normal actions. The cost of a virtual action is much greater than that of normal action. And it is set according to two variables: one is the maximum cost $C$ allowed by normal actions, and the other is the maximum number $N$ of actions allowed in the plan obtained by the planner. We uniformly set the costs of semi-virtual actions as $C \times N$, the costs of full virtual actions as $C \times N \times N$.

\begin{figure}[ht]
  \centering
  \includegraphics[width=4.8in]{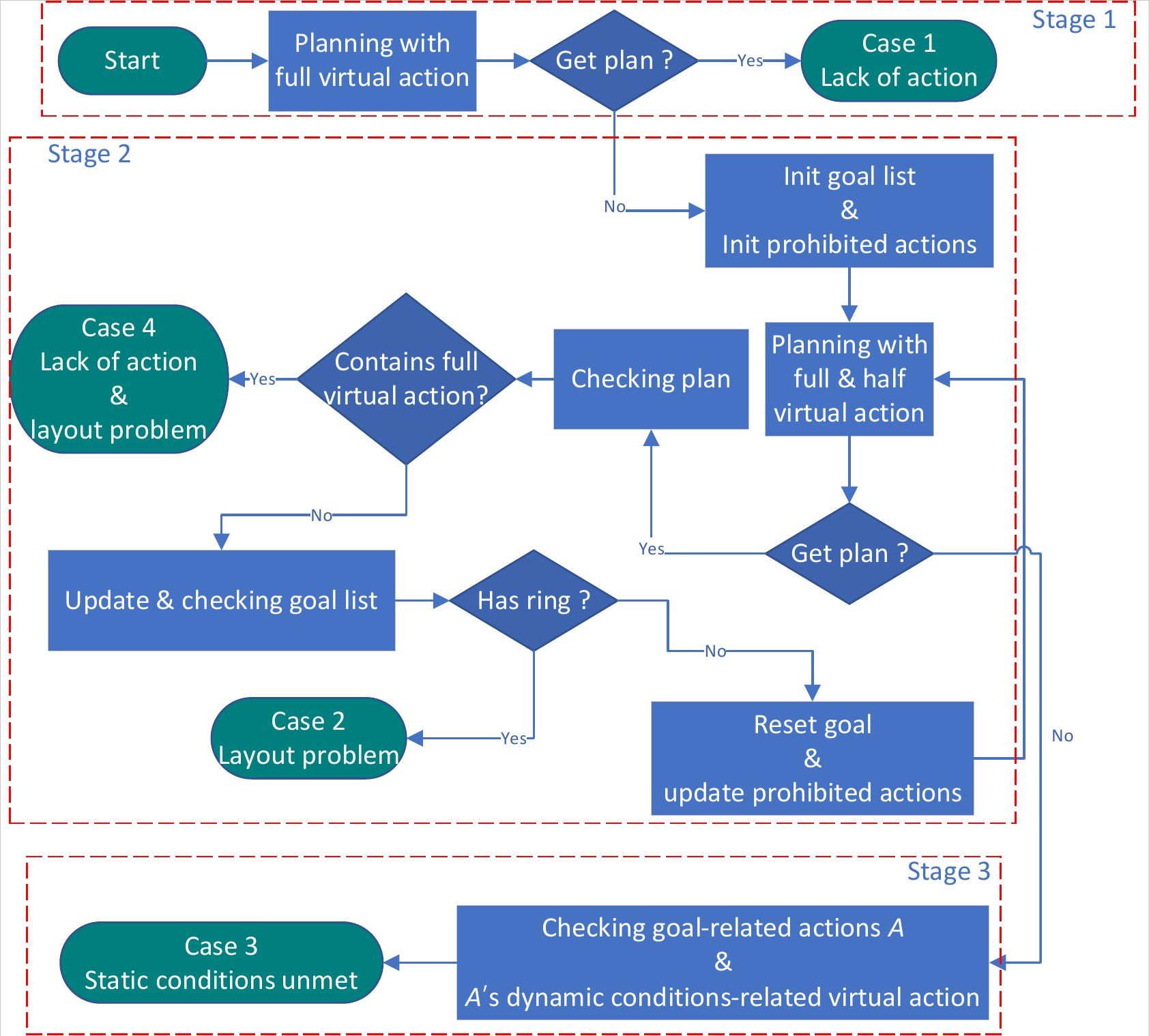}
  \caption{Framework overview.}
  \label{fig:framework}
\end{figure}

As Fig.\ref{fig:framework} shows, the planning algorithm starts after normal planning failed, can be divided into three stages. In the first stage, the action set used by the planner contains all full virtual actions, and if the planning fails, the planning algorithm enters the second stage. In the second stage, the action set also contains all semi-virtual actions, and the planning algorithm enters the third stage when the planner can't find a plan in the second stage.

In our method, a full virtual action is generated according to the dynamic predicate that is not the effect of any action, and a semi-virtual action is generated according to the dynamic predicate that is not the effect of any action. So, the plan generated in stage one must contain some full virtual actions, which represent a lack of normal actions to complete the corresponding state transitions. The reason may be design errors, or the robot lacks the corresponding capabilities.

In stage two, to determine if it is a layout problem, it is much more complex, and the steps are as follows:
\begin{enumerate}
  \item Initialize the goal list $L_g$, initially containing only the current goals $G$.
  \item Disable virtual actions related to $g$, that is, disable virtual actions with effect as $G$.
  \item Execute the planning, if the planner does not get any plan, go to step 7; if the plan obtained contains a full virtual action, enter step 6; if there is no full virtual action in the plan, but there is a semi-virtual action, go to step 4; if there is a ring in the list $L_g$, go to step 5.
  \item Take the effect of the first semi-virtual action in the plan as the new goal $g$, add it into $L_g$, enable the actions that were previously disabled and go to step 2.
  \item Case 2, there is a layout problem. We abandon to continue looking for other reasons and build the explanation according to the ring.
  \item Case 4, in this case, we abandon to continue looking for others reasons, build the explanation according to full Virtual actions.
  \item Go to stage 3.
\end{enumerate}

In stage three, two kinds of static conditions are not met, one is that \textbf{the static preconditions of the goal-related action} $A$ are not met, and the other is that \textbf{the preconditions of the virtual action related to the dynamic preconditions of $A$} are unmet. The former should be checked first, and the latter should be detected in the absence of abnormalities.

\subsection{Failure explanation}
In traditional planning algorithms, failed planning often returns no results, and no effective information can be obtained at this time. Our extended, progressive planning algorithm with virtual action can still return a result when the traditional planning fails. The full virtual actions or semi-virtual actions contained in the returned plan are used to explain the failure. Each virtual action has only the fewest preconditions and one effect. When it appears in the result, that means that its effect cannot be normally satisfied. A full virtual action means that the robot has no action to achieve its effect (case 1 \& case 4). For example, $full\_opened(D)$ appears in the plan, which means that the robot does not have the ability to open doors. A set of semi-virtual actions is used to locate the environmental factors. Their effects can be achieved through certain actions, but due to environmental constraints, the preconditions for the action to occur cannot be met (case 2). We use a simple example to illustrate this situation. The robot can open the door, but requires the room card, but the room card is locked in the room to be opened. In this way, a deadlock makes the robot unable to open the door. After our planning algorithm, we can get a ring to explain the failure. In addition, planning may fail because static conditions are not met, and no results are returned. However, after the previous process, it can be determined that the problem is that \textbf{the static conditions of the goal-related action} or \textbf{the static condition of the virtual actions related to the goal-related action's dynamic conditions} is unmet (case 3).

\section{Case study}\label{sec:cases}

\begin{figure}[ht]
  \centering
  \includegraphics[width=4.8in]{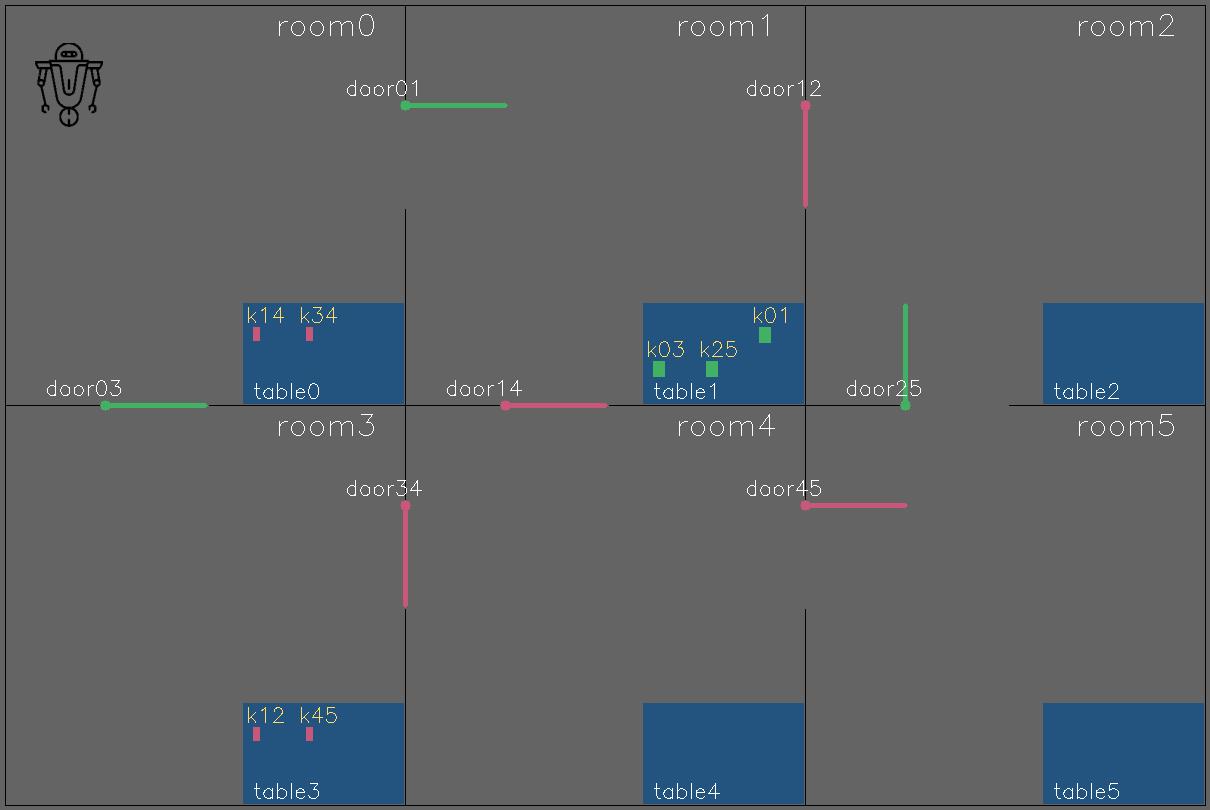}
  \caption{The world.}
  \label{fig:world}
\end{figure}

In this section, we will describe several type of case study. These cases cover lack of capabilities, layout problem, static condition unmet. Let's take the world shown in Figure~\ref{fig:world} as an example, there are a total of 6 rooms in the world, each room contains a table, each table can be placed some keys, each key can only be used to open the corresponding door. Initially the robot knows the location of all the keys as well as the state of all the doors, the robot is in $room0$, the goal is to move to $room5$. The static predicates included in the designed action domain are: $room(A)$, $table(A)$, $door(A)$, $key(A)$, $hand(A)$, $doorStatusValue(A)$, $handStatusValue(A)$, $inRoom(A,B)$, $connected(A,B,C)$, $isCard(A)$, $keyDoor(K,D)$. The predicate $doorStatusValue(A)$ means variable $A$ is a value of door status (closed or opened), $handStatusValue(A)$ means variable $A$ is a value of hand status (empty or full), $inRoom(A,B)$ means table $A$ in room $B$, $connected(A,B,C)$ means door $A$ connects room $B$ and $C$, $isCard(A)$ means key $A$ is a card, and $keyDoor(A,B)$ means key $A$ can used to open door $B$. The dynamic predicates in the action domain are shown in Table~\ref{tab:dynamic_predicate}.

\begin{table*}[ht!]
  \centering
  \caption{The dynamic predicates and their requirements.}
  \label{tab:dynamic_predicate}
  \begin{tabular}{|p{0.21\textwidth}|p{0.36\textwidth}|p{0.24\textwidth}|}
      \hline
      \textbf{Head} & \textbf{Requirements} & \textbf{Detail}\\ \hline
      \makecell*[lt]{$doorStatus(A,B)$} & \makecell*[lt]{$door(A)$ \& $doorStatusValue(B)$} & \makecell*[lt]{door is open or closed} \\ \hline
      \makecell*[lt]{$handStatus(A,B)$} & \makecell*[lt]{$hand(A)$ \& $handStatusValue(B)$} & \makecell*[lt]{hand is empty or full} \\ \hline
      \makecell*[lt]{$isPlaced(A,B)$} & \makecell*[lt]{$key(A)$ \& $table(B)$} & \makecell*[lt]{key is placed on table} \\ \hline
      \makecell*[lt]{$isHeld(A,B)$} & \makecell*[lt]{$key(A)$ \& $hand(B)$} & \makecell*[lt]{hand holds key}\\ \hline
      \makecell*[lt]{$robAt(A)$} & \makecell*[lt]{$room(A)$} & \makecell*[lt]{robot is in room $A$}\\
      \hline
  \end{tabular}
\end{table*}

We designed four actions for the robot: the robot moves into an adjacent room, picks up an object, puts down an object, and opens a door. The premise and effects of these actions are shown in Table~\ref{tab:action}.

\begin{table*}[ht!]
  \centering
  \caption{Action, its preconditions and effects.}
  \label{tab:action}
  \begin{tabular}{|p{0.30\textwidth}|p{0.275\textwidth}|p{0.28\textwidth}|}
    \hline
    \textbf{Action} & \textbf{Preconditions} & \textbf{Effects}\\ \hline
    \makecell*[lt]{$moveTo(R2,R1,D)$} & \makecell*[lt]{$robAt(R1)$ \\ \& $connected(D,R1,R2)$ \\ \& $doorStatus(D,opened)$} & \makecell*[lt]{$robAt(R2)$ \\ \& $\neg{robAt(R1)}$} \\ \hline
    \makecell*[lt]{$pickup(O,G,L,R)$} & \makecell*[lt]{$handStatus(G,empty)$ \\ \& $key(O)$ \\ \& $isPlaced(O,L)$ \\ \& $robAt(R)$ \\ \& $inRoom(L,R)$} & \makecell*[lt]{$\neg{handStatus(G,empty)}$ \\ \& $\neg{isPlaced(O,L)}$ \\ \& $isHeld(O,G)$} \\ \hline
    \makecell*[lt]{$putdown(O,G,L,R)$} & \makecell*[lt]{$robAt(R)$ \\ \& $isHeld(O,G)$ \\ \& $table(L)$ \\ \& $key(O)$ \\ \& $inRoom(L,R)$} & \makecell*[lt]{$isPlaced(O,L)$ \\ \& $handStatus(G,empty)$ \\ \& $\neg{isHeld(O,G)}$} \\ \hline
    \makecell*[lt]{$openDoor(D,G,K,R1,R2)$} & \makecell*[lt]{$robAt(R1)$ \\ \& $connected(D,R1,R2)$ \\ \& $keyDoor(K,D)$ \\ \& $isHeld(K,G)$ \\ \& $doorStatus(D,closed)$} & \makecell*[lt]{$\neg{doorStatus(D, closed)}$ \\ \& $doorStatus(D, opened)$}\\
    \hline
\end{tabular}
\end{table*}

\subsection{Case study 1: lack of action}
If the environment shown as Figure~\ref{fig:world} is the initial state, and the action domain does not contain action $openDoor(D,G,K,R1,R2)$ in Table~\ref{tab:action}. Planning with all normal actions, the planner can't find a way to $room5$. We can generate two full virtual actions that change the status of a door, as shown in Table~\ref{tab:full_action}.

\begin{table*}[ht!]
  \centering
  \caption{Full virtual action, its preconditions and effects.}
  \label{tab:full_action}
  \begin{tabular}{|p{0.30\textwidth}|p{0.275\textwidth}|p{0.28\textwidth}|}
    \hline
    \textbf{Action} & \textbf{Preconditions} & \textbf{Effects}\\ \hline
    \makecell*[lt]{$full\_e\_doorStatus(A,B)$} & \makecell*[lt]{$doorStatusValue(B)$ \\ \& $door(A)$ \\ \& not $doorStatus(A,B)$} & \makecell*[lt]{$doorStatus(A,B)$} \\ \hline
    \makecell*[lt]{$full\_d\_doorStatus(A,B)$} & \makecell*[lt]{$doorStatusValue(B)$ \\ \& $door(A)$ \\ \& $doorStatus(A,B)$} & \makecell*[lt]{$\neg{doorStatus(A,B)}$} \\
    \hline
\end{tabular}
\end{table*}

Planning with the actions added the full actions in Table~\ref{tab:full_action}, the planner can get a plan with four steps:
\begin{enumerate}
  \item $moveTo(room1,room0,door01)$;
  \item $full\_e\_doorStatus(door12,opened)$;
  \item $moveTo(room2,room1,door12)$;
  \item $moveTo(room5,room2,door25)$.
\end{enumerate}

There is a full virtual action $full\_e\_doorStatus(door12,opened)$ in the plan, we know that the normal actions do not contain an action to open the door. If the definition is forgotten, then this information can prompt the user; if the robot lacks this ability, it is also useful to know the specific information, such as explaining to the user or asking for help.

\subsection{Case2: the key is not reachable}

\begin{figure}[ht!]
  \centering
  \includegraphics[width=4.0in]{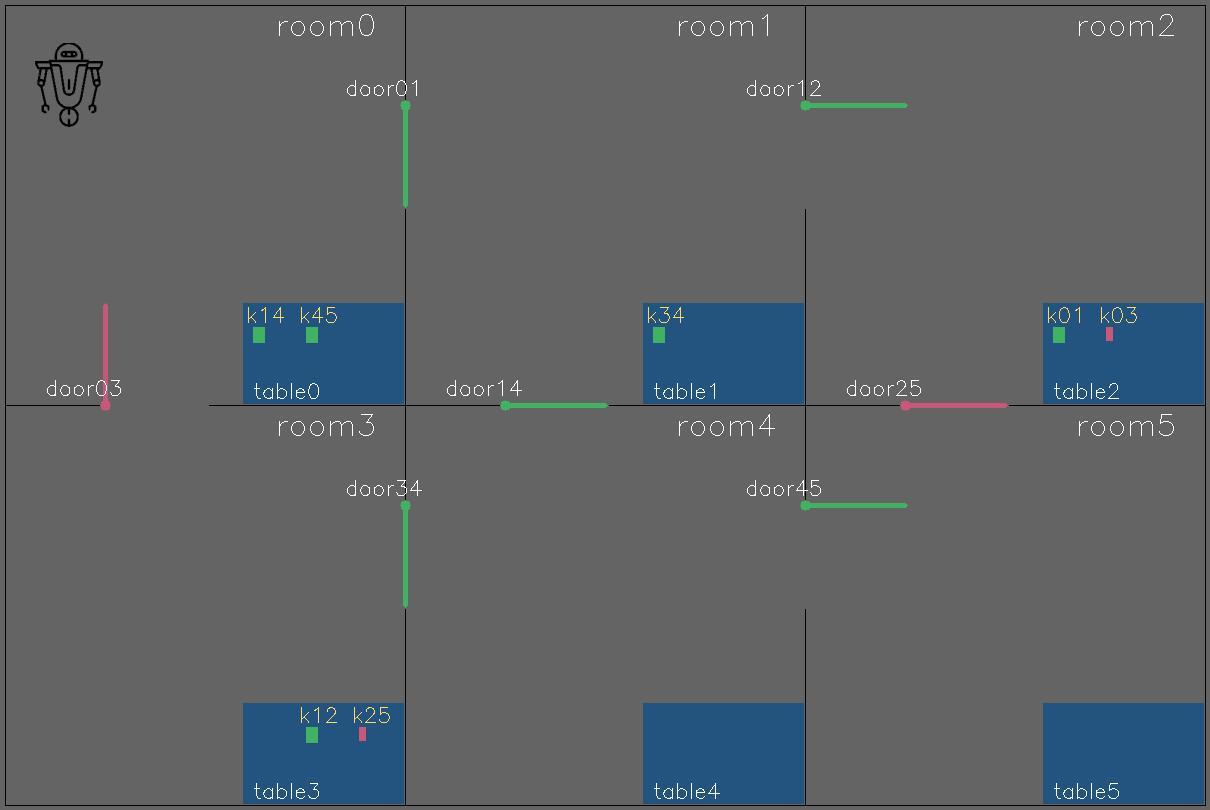}
  \caption{Case 2.}
  \label{fig:ring}
\end{figure}

Consider the environment shown in Figure~\ref{fig:ring} as the initial state, and the action set contains all the actions in Table~\ref{tab:action}. Planning with all normal actions, the planner can't find a way to $room5$. We can't generate any full virtual action, but can generate 10 semi-full actions as shown in Table~\ref{tab:semi_action}.

\begin{table*}[ht!]
  \centering
  \caption{Semi-virtual action, its preconditions and effects.}
  \label{tab:semi_action}
  \begin{tabular}{|p{0.30\textwidth}|p{0.275\textwidth}|p{0.28\textwidth}|}
    \hline
    \textbf{Action} & \textbf{Preconditions} & \textbf{Effects}\\ \hline
    \makecell*[lt]{$semi\_e\_handStatus(A,B)$} & \makecell*[lt]{$hand(A)$ \\ \& $handStatusValue(B)$ \\ \& not $handStatus(A,B)$} & \makecell*[lt]{$handStatus(A,B)$} \\ \hline
    \makecell*[lt]{$semi\_d\_handStatus(A,B)$} & \makecell*[lt]{$hand(A)$ \\ \& $handStatusValue(B)$ \\ \& $handStatus(A,B)$} & \makecell*[lt]{$\neg{handStatus(A,B)}$} \\ \hline
    \makecell*[lt]{$semi\_e\_isHeld(A,B)$} & \makecell*[lt]{$hand(B)$ \& $key(A)$ \\ \& not $isHeld(A,B)$} & \makecell*[lt]{$isHeld(A,B)$} \\ \hline
    \makecell*[lt]{$semi\_d\_isHeld(A,B)$} & \makecell*[lt]{$hand(B)$ \& $key(A)$ \\ \& $isHeld(A,B)$} & \makecell*[lt]{$\neg{isHeld(A,B)}$} \\ \hline
    \makecell*[lt]{$semi\_e\_doorStatus(A,B)$} & \makecell*[lt]{$doorStatusValue(B)$ \\ \& $door(A)$ \\ \& not $doorStatus(A,B)$} & \makecell*[lt]{$doorStatus(A,B)$} \\ \hline
    \makecell*[lt]{$semi\_d\_doorStatus(A,B)$} & \makecell*[lt]{$doorStatusValue(B)$ \\ \& $door(A)$ \\ \& $doorStatus(A,B)$} & \makecell*[lt]{$\neg{doorStatus(A,B)}$} \\ \hline
    \makecell*[lt]{$semi\_e\_isPlaced(A,B)$} & \makecell*[lt]{$table(B)$ \& $key(A)$ \\ \& not $isPlaced(A,B)$} & \makecell*[lt]{$isPlaced(A,B)$} \\ \hline
    \makecell*[lt]{$semi\_d\_isPlaced(A,B)$} & \makecell*[lt]{$table(B)$ \& $key(A)$ \\ \& $isPlaced(A,B)$} & \makecell*[lt]{$\neg{isPlaced(A,B)}$} \\ \hline
    \makecell*[lt]{$semi\_e\_robAt(A)$} & \makecell*[lt]{$room(A)$ \\ \& not $robAt(A)$} & \makecell*[lt]{$robAt(A)$} \\ \hline
    \makecell*[lt]{$semi\_d\_robAt(A)$} & \makecell*[lt]{$room(A)$ \\ \& $robAt(A)$} & \makecell*[lt]{$\neg{robAt(A)}$} \\ 
    \hline
\end{tabular}
\end{table*}

\begin{figure}[ht]
  \centering
  \includegraphics[width=4in]{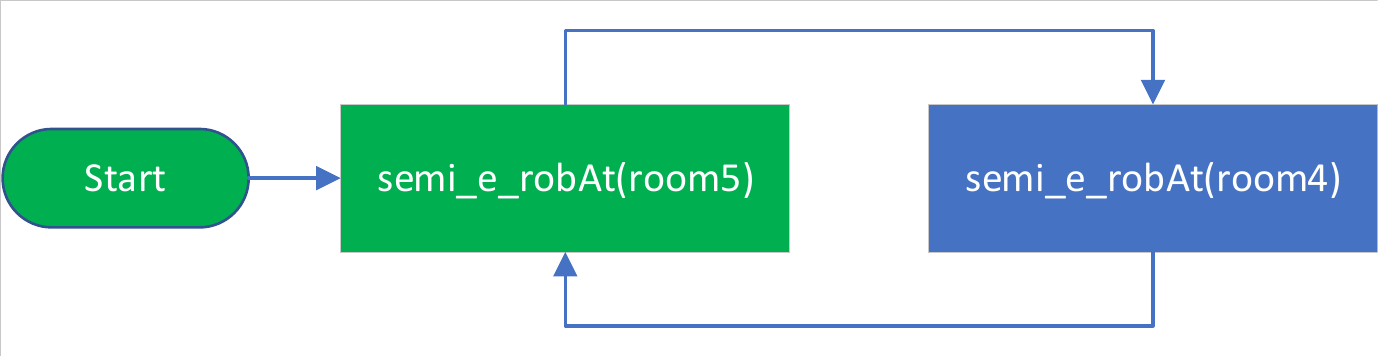}
  \caption{The goal list.}
  \label{fig:ring1}
\end{figure}

Planning with the actions added the full actions in Table~\ref{tab:full_action}, the planner can get a goal list with three semi-virtual actions as Figure~\ref{fig:ring1}

\begin{figure}[ht]
  \centering
  \includegraphics[width=4in]{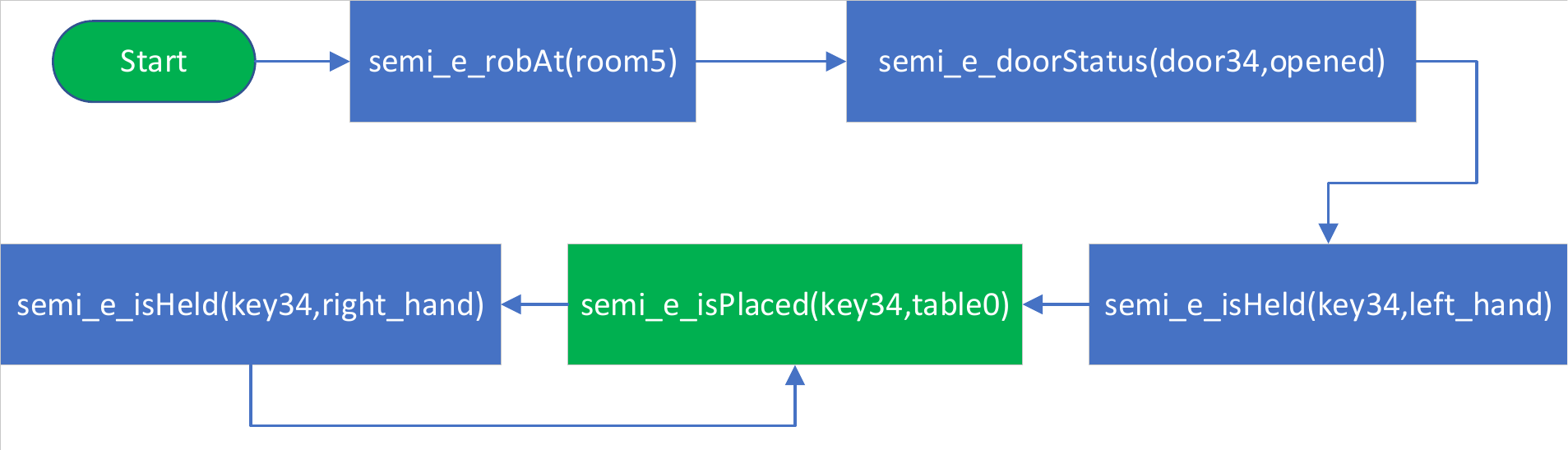}
  \caption{The goal list without semi-virtual action related to $robAt(A)$.}
  \label{fig:ring2}
\end{figure}

The goal list contains a ring. It means: to get to $room5$, the robot needs to reach $room4$ first, and to reach $room4$, it needs to reach $room5$ first. There is little valid information about this ring. In fact, in the action domain of this experiment, the position of the robot itself provides important order and logical limitations, so the semi-virtual action of changing its position should not be established. It is possible only if the dynamic predicate $robAt(A)$, which represents the robot's position,  has no related semi-virtual action. Because the conditions of its related normal action can be achieved by other virtual actions. Remove the last two semi-full actions in Table~\ref{tab:semi_action} and the planner can get a goal list as Figure~\ref{fig:ring2}. From Figure~\ref{fig:ring2} we know that opening the door needs to hold $key34$, but the position of $key34$ is unreachable.

\subsection{Case 3: card or not}

\begin{figure}[ht]
  \centering
  \includegraphics[width=4in]{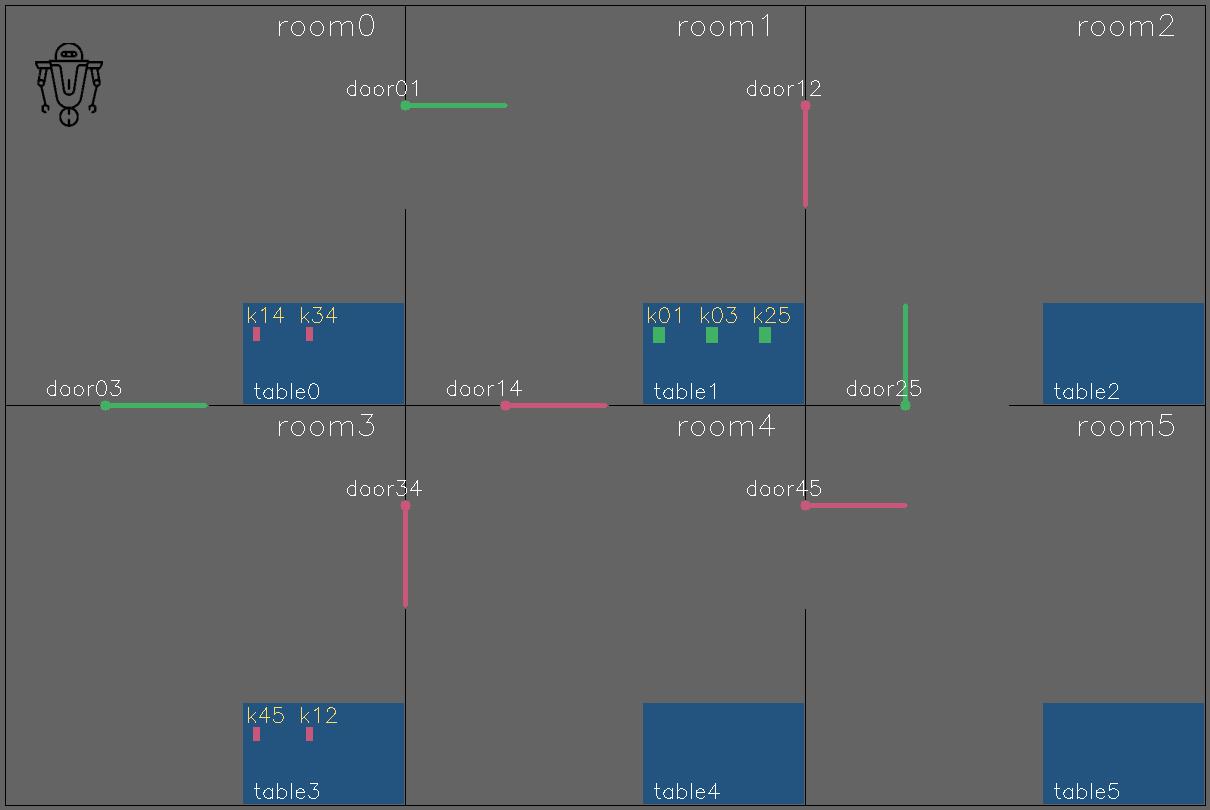}
  \caption{Case 3.}
  \label{fig:static}
\end{figure}

In this case, the initial state is shown in Figure~\ref{fig:static}, the red keys on the tables are not the card, the green keys on the tables are cards. The action set contain the actions of Table~\ref{tab:action} and Table~\ref{tab:semi_action} (except the last two semi-virtual actions). We modify the Actions $pickup(O,G,L,R)$, $semi\_e\_isHeld(A,B)$ and $semi\_d\_isHeld(A,B)$ as shown in Table~\ref{tab:pickup}. The new action defines that only when a key is a card it can be grasped by the robot.

\begin{table*}[ht!]
  \centering
  \caption{Action, its preconditions and effects.}
  \label{tab:pickup}
  \begin{tabular}{|p{0.30\textwidth}|p{0.275\textwidth}|p{0.28\textwidth}|}
    \hline
    \textbf{Action} & \textbf{Preconditions} & \textbf{Effects}\\ \hline
    \makecell*[lt]{$pickup(O,G,L,R)$} & \makecell*[lt]{$handStatus(G,empty)$ \\ \& $key(O)$ \\ \& \textbf{\textit{isCard(O)}} \\ \& $isPlaced(O,L)$ \\ \& $robAt(R)$ \\ \& $inRoom(L,R)$} & \makecell*[lt]{$\neg{handStatus(G,empty)}$ \\ \& $\neg{isPlaced(O,L)}$ \\ \& $isHeld(O,G)$} \\ \hline
    \makecell*[lt]{$semi\_e\_isHeld(A,B)$} & \makecell*[lt]{$hand(B)$ \& $key(A)$ \\ \& not $isHeld(A,B)$ \\ \& \textbf{\textit{isCard(A)}}} & \makecell*[lt]{$isHeld(A,B)$} \\ \hline
    \makecell*[lt]{$semi\_d\_isHeld(A,B)$} & \makecell*[lt]{$hand(B)$ \& $key(A)$ \\ \& $isHeld(A,B)$ \\ \& \textbf{\textit{isCard(A)}}} & \makecell*[lt]{$\neg{isHeld(A,B)}$} \\
    \hline
\end{tabular}
\end{table*}

Planning with the actions, the planner can get a goal list with two semi-virtual actions: $semi\_e\_robAt(room5)$, $semi\_e\_doorStatus(door14,opened)$. We get a planning failure, then from $semi\_e\_doorStatus(door14,opened)$, we find related action $openDoor(door14,G,K,R1,R2)$, checks its static conditions, that can be satisfied. $G$ can be $left\_hand$ or $right\_hand$, $K$ is $key14$, $(R1,R2)$ can be $(room1,room4)$ or $(room14,room1)$. Table~\ref{tab:dsa} lists the dynamic conditions of $openDoor$, the actions related to the dynamic conditions, and these actions' static conditions. All the static conditions can be met except $isCard(key14)$, $key14$ is not a card. We report this information: the robot can not pickup the $key14$ due to it is not a card, but the robot need to open the door with $key14$.

\begin{table*}[ht!]
  \centering
  \caption{Dynamic condition, action, and static conditions}
  \label{tab:dsa}
  \begin{tabular}{|p{0.30\textwidth}|p{0.275\textwidth}|p{0.28\textwidth}|}
    \hline
    \textbf{Dynamic conbditions} & \textbf{Action} & \textbf{Static Conditions}\\ \hline
    \makecell*[lt]{$robAt(R1)$} & \makecell*[lt]{$moveTo(R1,R0,D0)$} & \makecell*[lt]{$connected(D0,R0,R1)$} \\ \hline
    \makecell*[lt]{$isHeld(K,G)$} & \makecell*[lt]{$pickup(K,G,L0,R0)$} & \makecell*[lt]{$key(K)$ \\ \& $isCard(K)$} \\ \hline
    \makecell*[lt]{$isHeld(K,G)$} & \makecell*[lt]{$semi\_e\_isHeld(K,G)$} & \makecell*[lt]{$key(K)$ \\ \& $isCard(K)$ \\ \& $hand(G)$} \\
    \hline
\end{tabular}
\end{table*}

\section{Conclusion}\label{sec:conclusion}
This paper presents a planning framework with virtual actions for addressing the reason for task planning failure. Full virtual action, semi-virtual action, and static precondition detection are used to determine whether precondition of the action are satisfied: 1) the state that is not an effect of any action; 2) the state that is an effect of some action, and 3) the static preconditions of an action. We set different costs for normal actions, semi-virtual actions, and full virtual actions, the former being much smaller than the latter, so that the planner takes precedence over normal actions, followed by semi-virtual actions, then full virtual actions. The information about the failure can assist the user in debugging the design of the action models, and it is also useful in requesting the assistance of other agents. In the future, we will try to improve planning efficiency.

\begin{acknowledgements}
This research was funded by the National Natural Science Foundation of China grant number U1613216,61573333.
\end{acknowledgements}

%
%
The authors declare that they have no conflict of interest.

\bibliographystyle{spmpsci}      
\bibliography{planning.bib}   


\end{document}